# Intrinsically motivated reinforcement learning for human-robot interaction in the real-world

Ahmed H. Qureshi*, Yutaka Nakamura, Yuichiro Yoshikawa, Hiroshi Ishiguro

*Department of System Innovation, Graduate School of Engineering Science, Osaka University, 1-3 Machikaneyama, Toyonaka, Osaka, Japan.*

**Abstract**

For a natural social human-robot interaction, it is essential for a robot to learn the human-like social skills. However, learning such skills is notoriously hard due to the limited availability of direct instructions from people to teach a robot. In this paper, we propose an intrinsically motivated reinforcement learning framework in which an agent gets the intrinsic motivation-based rewards through the action-conditional predictive model. By using the proposed method, the robot learned the social skills from the human-robot interaction experiences gathered in the real uncontrolled environments. The results indicate that the robot not only acquired human-like social skills but also took more human-like decisions, on a test dataset, than a robot which received direct rewards for the task achievement.

## 1. Introduction

The field of Human-Robot Interaction (HRI) has emerged with an objective of socializing robots [1]. One of the biggest challenges faced by sociable robots is the challenge of interpreting complex human interactive behaviors [2]. These interactive behaviors are not only complex because of their enormous diversity

---

*Corresponding author

 *Email address:* qureshi.ahmed@irl.sys.es.osaka-u.ac.jp (Ahmed H. Qureshi)

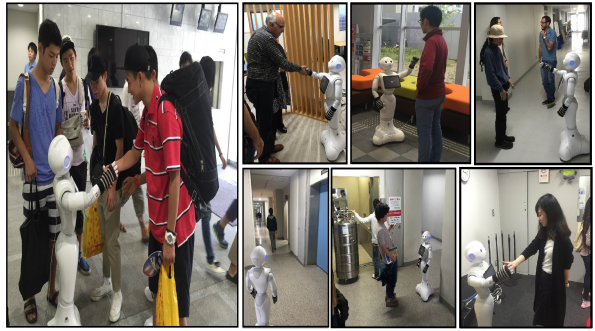

Figure 1: Robot learning social interaction skills from people.

but also because of being driven by the people's underlying invisible intentions and social norms [2] [3]. It is therefore nearly impossible to devise a hand-crafted policy for the robots to interact with people. Perhaps, the only practical way to realize social HRI is to build robots that can learn autonomously [4] through interaction with people in the real world.

Recently, the notion of building robots that learn through interaction with people has gained the interest of many researchers and thus, has led to the proposition of various approaches. The notable methods include [5] [6] [7]. In [5], the robot learned to choose appropriate responsive behavior based on the human intentions inferred from their body movements. However, we believe that the human intentions are not only driven by the body movements but also by other intention depicting factors such as gaze directions, face expression, body language, walking trajectories, on-going activities, and surrounding environment. In [6] [7], the interaction between two persons is recorded to train a robot to respond to a human partner by imitating the behavior of another human interacting partner from the recorded dataset. However, in all methods mentioned above, only a single interaction partner is considered for the interaction with a robot whereas, in the real-world, the robot can be approached by any number of people. Hence to cope with the problem of learning human-like interaction skills, it is important for a robot to learn through interaction with people in their natural and uncontrolled ambient.

Reinforcement learning (RL) attempts to solve the challenge of building a robot that learns through interaction (i.e., taking actions) with the physical world [8]. However, the difficulties in the perception tasks were hampering the field of end-to-end reinforcement learning until the recent development of Deep Q-Network (DQN) [9]. The DQN combines deep learning [10] with reinforcement learning [8], and it learned to map high-dimensional images to actions for playing various arcade Atari games. The idea of deep Q-learning (or DQN) has recently been extended to the robotics field. For instance, in [11], twenty continuous control problems including legged locomotion, car driving, and cart-pole swing-up, in a simulated environment, were addressed through Deep Reinforcement Learning. Likewise, in [12], deep visuomotor policies were learned through guided policy search to perform various tasks such as screwing a cap on a bottle, and peg insertion. More recently, we extended the application of DQN from simulated environments to the problem of a robot learning social interaction skills through interaction with people in open public places (e.g., cafeteria, department reception, common rooms, etc.) [13], [14].

In [13], the robot acquired social intelligence after 14 days of interaction with people in free public places (see Fig. 1). The results indicate the success of [13] in interpreting complex human behaviors and choosing the appropriate actions for the interaction. Furthermore, the method in [14], extends [13] by adding a neural attention model to it. The attention model enabled the robot to indicate its attention thus leading to perceivable social human-robot interaction.

However, one of the limitations of [13] [14], inherited from RL, is the requirement of an external reward for the learning. This external reward is given to the RL-agent on the task accomplishment. In HRI, the direct rewards are usually scarce which makes the learning difficult. In this paper, we propose an intrinsically motivated deep reinforcement learning framework for learning the social interaction skills in the real reward-sparse world.

The proposed method comprises an action-conditional prediction network (Pnet) and a policy network (Qnet). The Qnet is intrinsically motivated by Pnet for learning the social interaction skills. We consider the same problem setting as for [13], i.e., the robot learns to greet people in public places with a set of four actions (wait, look towards human, wave hand and handshake). As the problem setting of [13] and the proposed work is same, we exploit the human-robot interaction experiences collected in [13] for the off-policy training of our intrinsically motivated reinforcement learning framework. Moreover, we also evaluate the performance of both Qnet and Pnet on a test dataset (not seen by the system during training) as well as through experimentation in the real-world. Also, the source code and the dataset of this work will be made available online[1].

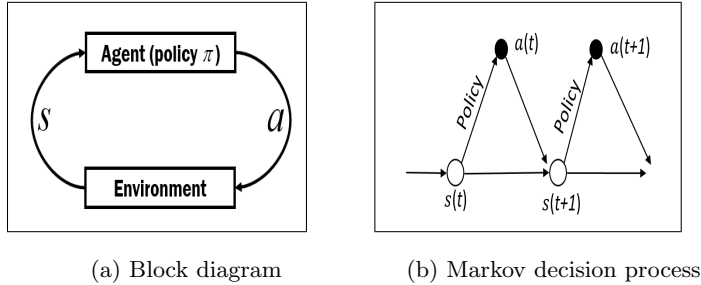

(a) Block diagram

(b) Markov decision process

Figure 2: Reinforcement learning

The remainder of the paper is organized as follows. Section 2 provides the problem formulation and the previous methods of modeling the intrinsic motivation and their applications in robotics. Section 3 and 4 explain the proposed model and its implementation details, respectively. Section 5 presents the results while Section 6 provides the discussion on the results. Finally, Section 7 concludes the paper together with some pointers to the future areas of research.

## 2. Preliminaries

This section formulates the Q-learning framework [8] which is a well-known Reinforcement Learning approach. It also describes the existing intrinsic moti-vation models and their robotics applications.

---

[1]https://sites.google.com/a/irl.sys.es.osaka-u.ac.jp/member/home/ahmed-qureshi/deephri

We consider a discrete time Markov decision process, as shown in Fig. 2, in which the agent interacts with an environment. At time $t$, the agent observes a state $s(t)$ and takes a discrete action $a(t)$ from a finite set of legal actions $\mathcal{A} = \{1, 2, \ldots, |\mathcal{A}|\}$ under a policy $\pi$. This action execution leads to a state transition from $s(t)$ to $s(t + 1)$ and a scalar reward $r(s(t), a(t)) \in \mathbb{R}$. The ultimate goal of the agent is to learn a policy $\pi$ that maximizes the expected total return,

$$R\big(s(t), a(t)\big) = \mathbb{E}\bigg[\sum_{t'=t}^{T} \gamma^{t'-t} r\big(s(t'), a(t')\big)\bigg], \tag{1}$$

where $T$ and $\gamma \in (0, 1]$ correspond to the terminal time and discount factor, respectively. For rest of the paper, we denote reward $r(s(t), a(t))$ as $r(t)$ for brevity.

### 2.1. Q-learning

In the Q-learning [8] [15], the policy $\pi$ is determined from an *action-value* function $Q(s, a)$ which indicates the expected return when the agent is in state $s$ and takes an action $a$. Furthermore, the optimal *action-value* function represents the expected return for taking an action $a$ in a state $s$ and thereafter, following an optimal policy $\pi$ i.e.,

$$Q^*(s, a) = \max_{\pi} \mathbb{E}\bigg[R\big(s(t), a(t)\big)\big|s(t) = s, a(t) = a, \pi\bigg] \tag{2}$$

This optimal *action-value* function also obeys a recursive Bellman relation which is formalized as follows:

$$Q^*(s, a) = \mathbb{E}\bigg[r(t) + \gamma \max_{a'} Q^*\big(s(t+1), a'\big)\big|s(t) = s, a(t) = a\bigg] \tag{3}$$

In the general RL frameworks, the reward $r$ is assumed to come from an external environment. However, due to the lack of availability of an external reward signal in the interactive learning scenarios, the researchers have come up with the notion of intrinsic motivation. The following sections describe the previously proposed state-of-the-art intrinsic motivation models and their explored applications in the robotics.

## 2.2. Intrinsic motivations

This section provides a brief overview of state-of-the-art intrinsic motivation models that have been previously proposed under the framework of reinforcement learning.

### 2.2.1. Competence-based motivation

These models have two levels of action. The first level decides what goals should be explored for reaching and the second level decides what action should be taken to reach those goals (for details see [16]). In general, the reward value is determined by measuring the extent to which the agent has reached the self-determined goal.

### 2.2.2. Novelty-based motivation

These models provide a higher reward to the agent on seeing a novel or new information [16]. For example in predictive novelty motivation models [17], the rewards are directly proportional to the prediction errors. The novelty-based models are also sometimes known as curiosity-driven methods as they make an agent curious to acquire new knowledge (see [18] for an in-depth discussion of these models).

### 2.2.3. Empowerment

The empowerment based rewarding directs an agent to take a sequence of actions that can bring maximum information to the agent itself. Recently, Shakir et. al [19] developed a method for approximating empowerment. They employ variational methods to approximate the empowerment between an agents actions and the future state of the environment. However, the effectiveness of their approach has only been demonstrated in simulated static and dynamic environments.

### 2.2.4. Surprise-based motivation

These methods give a reward on surprise, i.e., an agent gets excited in the situations where its understanding of the world goes wrong. These models are often confused with novelty-based intrinsic motivation methods; however, there is a distinction between them. In novelty-based methods, the agent gets excited on seeing a new information whereas in surprise-based methods, the agent gets higher reward when its prediction of future states goes wrong (for details see [20]). Recently, a deep neural networks based method has also been proposed for modeling the surprise-based motivation [21]. However, the performance of this method has only been demonstrated in the animated world such as Atari games instead of real world problems.

### 2.2.5. Learning progress based motivation

These models encourage the agent to improve its understanding of the world. For example, in prediction-based intrinsic motivation models, the learning progress-based motivation could be to minimize the prediction errors over the time. A notable work in this direction is the work by Oudeyer et. al [22]. However, the applicability of this method is limited to simple toy problems.

### 2.3. Robotics applications of Intrinsically Motivated RL

To date, the applicability of intrinsically motivated reinforcement learning methods is limited to simple problems [23]. For instance, Stout and Barto [24] proposed competence-based intrinsic motivation for skills acquisition and demonstrated its application in simple and artificial 2D grid-worlds. Likewise, Pape et al. [25] exploit curiosity-driven reinforcement learning for the acquisition of tactile skills on a biomimetic robotic finger. In a similar vein, Ngo et al. [26] [27] investigated reinforcement learning system driven by progress-based curiosity for performing simple tasks such as placing an object.

In the intrinsic motivation based learning methods presented so far in this section, the direct reward signal is usually augmented with the intrinsic motivation to promote exploration. However, in our proposed method, the predictive model based intrinsic motivation solely lays down the learning objective for the policy network. The reason for considering only the intrinsic motivation as a learning objective, in our proposed work, is the assumption that human behav-ior is predictable. Hence, the predictive intrinsic motivation model is assumed to be sufficient for the real world HRI applications.

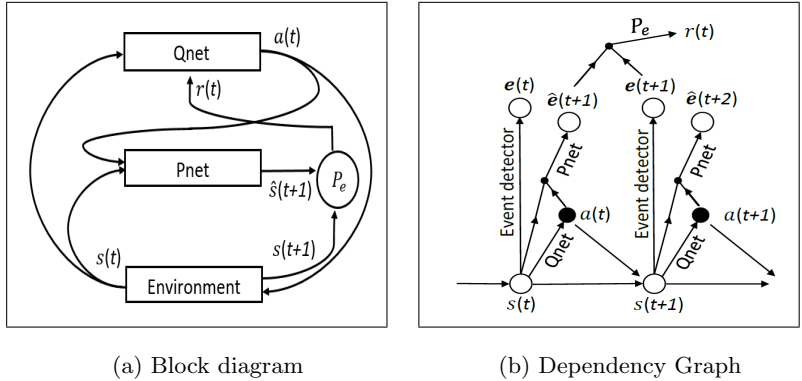

(a) Block diagram    (b) Dependency Graph

Figure 3: Intrinsically motivated deep reinforcement learning

Furthermore, unlike previously proposed methods, we demonstrate the effectiveness of our novel approach in a high-dimensional complex human-robot interaction problem in which a robot needs to learn the social interaction skills through interaction with people in open public places such as a cafeteria, common rooms, and department receptions (see Fig. 1).

## 3. The Proposed Method

This section presents our proposed method which includes the Qnet, Pnet and their training procedure.

In each time step, several events which are considered to be relevant to the task are detected by an event detector (Fig. 3(b)). For a robot to learn the social interaction skills, we consider 3 events, i.e., handshake, eye contact and smile. We assume that these events can be detected using the existing techniques such as OpenCV based methods. For the current implementation of the event detector, please refer to the section 4.2. The $\boldsymbol{e}(t)$ in Fig. 3(b) denotes the occurrences of the events at time $t$, and each element in $\boldsymbol{e}(t)$ corresponds to one of the events. For example, $e_j(t) = 1$ indicates the $j$-th event occurs while $e_j(t) = 0$ indicates it does not occur.

Overall, Qnet chooses an action in a given state, and Pnet predicts the occurrences of events $\hat{e}(t+1)$ at the next time step according to the state-action pair (see Fig. 3). The internal reward, at time $t$, for Qnet is determined according to the prediction error ($P_e$) between Pnet's prediction $\hat{e}(t+1)$ and the actual occurrence of events $\boldsymbol{e}(t+1)$.

### 3.1. Policy network (Qnet)

Qnet is a deep neural networks based *action-value* function approximator. The input to this function is a state $s$. In this research, a tuple of observed images is treated as a state and thus, the input becomes a high-dimensional vector consisting of pixel values of all images. The output of the network is a $K$-dimensional vector and the $k$-th element corresponds to the value of the $k$-th action.

Qnet is trained so that each output value is adjusted towards the expected return (Eq.(1)), i.e.,

$$q_k(s) \approx R(s, a_k), \tag{4}$$

where $q_k(s)$ is a value of $k$-th element of an output vector corresponding to the $k$-th action. The training objective for Qnet is as follows:

$$L(\boldsymbol{\theta}) = \mathbb{E}_{(s,a,r,s') \sim \mathcal{M}} \left[ \left( B - Q(s, a; \boldsymbol{\theta}) \right)^2 \right], \tag{5}$$

where $B$ corresponds to the Bellman target, i.e., $B = r + \gamma \max_{a'} Q(s', a'; \hat{\boldsymbol{\theta}})$. The functions $Q(s', a'; \hat{\boldsymbol{\theta}})$ and $Q(s, a; \boldsymbol{\theta})$ represent the target Q-network with old parameters $\hat{\boldsymbol{\theta}}$, and learning Q-network with recently updated parameters $\boldsymbol{\theta}$, respectively. The old parameters $\hat{\boldsymbol{\theta}}$ are updated to current parameters $\boldsymbol{\theta}$ after every C-steps (see Algorithm 1).

The gradient of the above-mentioned objective function takes the following form:

$$\nabla_{\theta_i} L(\boldsymbol{\theta}) = \mathbb{E}_{s,a,r,s'} \left[ \left( B - Q(s, a; \boldsymbol{\theta}) \right) \nabla_{\theta_i} Q(s, a; \boldsymbol{\theta}) \right] \tag{6}$$

This gradient is used to update the learning parameters $\theta_i$. Furthermore, like DQN [9], we use experience replay [28] to train Qnet. Hence, the above-mentioned equations, 5 and 6, are computed using the batch of interaction experiences $(s, a, r, s')$ sampled randomly from a replay memory $\mathcal{M}$.

The structure of our proposed Qnet is shown in Fig. 4(b). It is a dual stream deep convolutional neural network which maps pixels (high-dimensional states) to $q$-values of the legal actions. The two streams, grayscale channel and depth channel, process the grayscale and depth images, respectively. The output from these two streams are only fused together for taking a greedy action (action with maximum $q$-value). This fusion is performed by normalizing the $q$-values from both streams and then taking their average.

### 3.2. Action-conditional Prediction network (Pnet)

Pnet is an event predictor which is implemented as a multi-label classifier (see Fig. 4(a)), and it is formalized as follows:

$$\hat{\boldsymbol{e}}(t+1) = \text{Pnet}\big(s(t), a(t)\big), \tag{7}$$

where $s$, $a$ and $\hat{\boldsymbol{e}}$ are the state, action and occurrence probabilities of events, respectively. The $j$-th element of $\hat{\boldsymbol{e}}(t+1)$ indicates the occurrence probability of an event $j$ at the next step. Furthermore, Pnet is trained to minimize the prediction error $\hat{\boldsymbol{e}} - \boldsymbol{e}$, and to do so, we employ the following Binary Cross Entropy (BCE) loss function:

$$\text{BCE}(\boldsymbol{e}, \hat{\boldsymbol{e}}) = -\frac{1}{J}\sum_{i=1}^{J}(e_i \cdot \log(\hat{e}_i) + (1 - e_i) \cdot \log(1 - \hat{e}_i)), \tag{8}$$

where $J$, $\hat{\boldsymbol{e}}$ and $\boldsymbol{e}$ correspond to the number of events (in this case $J = 3$) related to the task, output of Pnet and output of the event detector, respectively.

The reward function $g : \hat{\boldsymbol{e}}, \boldsymbol{e} \to r$ then computes the reward $r$ on the basis of the prediction error between the predicted event, $\hat{\boldsymbol{e}}$, and the event, $\boldsymbol{e}$, that actually happened.

### 3.3. Learning procedure

Algorithm 1 outlines our proposed method for a robot to learn social interaction skills from people in the uncontrolled environments. The proposed training

---

**Algorithm 1:** Learning the social interaction skills

---

**1** Initialize replay memory $\mathcal{M}$ to size $N$

**2** Initialize Pnet$(s, a; \boldsymbol{\theta}')$ with parameters $\boldsymbol{\theta}'$

**3** Initialize the learning Qnet $Q(s, a; \boldsymbol{\theta})$ with parameters $\boldsymbol{\theta}$

**4** Initialize the target Qnet $Q(s, a; \hat{\boldsymbol{\theta}})$ with parameters $\hat{\boldsymbol{\theta}} = \boldsymbol{\theta}$

**5** **for** episode $= 1, M$ **do**

**6**   **Data generation phase:**

**7**   Initialize the start state to $s_1$

**8**   **for** $i = 1, T$ **do**

**9**    With probability $\epsilon$ select a random action $a(t)$ otherwise select
   $a(t) = \max_{\mathrm{a}} Q(s(t), \mathrm{a}; \boldsymbol{\theta})$

**10**    $s(t+1) \leftarrow \text{ExecuteAction}\big(a(t)\big)$

**11**    $\boldsymbol{e}(t+1) \leftarrow \text{EventDetector}\big(s(t+1)\big)$

**12**    Store the transition $\big(s(t), a(t), \boldsymbol{e}(t+1), s(t+1)\big)$ in $\mathcal{M}$

**13**   **Learning phase:**

**14**   Randomize a memory $\mathcal{M}$ for experience replay

**15**   **for** $i = 1, n$ **do**

**16**    Sample a random minibuffer $\mathcal{B}$ from $\mathcal{M}$

**17**    **while** $\mathcal{B}$ **do**

**18**     Sample a minibatch $m$ of transitions $(s_k, a_k, \boldsymbol{e}_{k+1}, s_{k+1})$ from
    $\mathcal{B}$ without replacement

**19**     Perform SGD on BCE loss w.r.t Pnet parameters $\boldsymbol{\theta}'$

**20**     $r = \text{ComputeRewards}(m, \text{Pnet})$

**21**     Compute Bellman targets:

$$B_k = \begin{cases} r_k, \text{if step } k{+}1 \text{ is terminal} \\ r_k + \gamma \max_{\mathrm{a}} Q(s_{k+1}, \mathrm{a}; \hat{\boldsymbol{\theta}}), \text{otherwise} \end{cases}$$

**22**     Perform gradient descent on loss $(B_k - Q(s_k, a_k; \boldsymbol{\theta}))^2$ w.r.t
    the Qnet parameters $\boldsymbol{\theta}$

**23**   After every C-episodes sync $\hat{\boldsymbol{\theta}}$ with $\boldsymbol{\theta}$.

---

procedure comprises of two phases, i.e., data generation phase and learning phase.

In data generation phase, the robot interacts with an environment using $\epsilon$-greedy policy. For non-greedy policy, an action is picked randomly from $\mathcal{A}$ whereas to take the greedy action, the learning Qnet $Q(s, a; \boldsymbol{\theta})$ is used. Finally, the interaction experiences gathered in this phase are stored into a replay memory $\mathcal{M}$.

In learning phase, the updated replay memory $\mathcal{M}$ is used to train Pnet followed by the training of Qnet. Note that both streams of Qnet are trained independently by minimizing the equation 5. Although we use a standard Q-learning approach (as in [9]) to optimize Qnet, any advance optimization technique such as TRPO [29] and PPO [30] can be applied since these methods only require that the policy be differentiable. Furthermore, investigation of using the different policy optimization methods on Qnet's performance remains a part of our future studies.

## 4. Implementation details

This section gives the implementation details of the proposed project[2]. The proposed neural models, Pnet and Qnet, are implemented in torch/lua and the robot side programming is done in python. The system used for training has 3.40GHz$\times$8 Intel Core i7 processor with 32 GB RAM and GeForce GTX 980 GPU. The remaining section explains different modules of the project.

### 4.1. Robotic system

Aldebaran's Pepper[3] robot was used to gather the interaction experiences. The built-in 2D camera on Pepper's forehead and a 3D sensor behind the Pepper's right eye were used to obtain the grayscale and depth images, respectively.

---

[2]The source code and the interaction experiences are available at https://sites.google.com/a/irl.sys.es.osaka-u.ac.jp/member/home/ahmed-qureshi/deephri

[3]https://www.ald.softbankrobotics.com/en/cool-robots/pepper

Both visual sensors were operated at 10 frames per second with a resolution of 320 × 240. An additional FSR touch sensor was pasted on the robot's right hand. This touch sensor was covered under the gloves for aesthetic reasons as shown in Fig. 1.

The robot could only interact with the environment using four actions, i.e., wait, look towards human, wave hand and handshake. During the actions other than waiting, the robot becomes sensitive to sound and movement stimulus. This sensitivity implies that if the robot senses any of the stimuli, it looks at the stimulus origin to check for the presence of a person. In case any person is detected at that origin, the robot tracks them with its head. The description of the four actions is as follows:

*1) Wait (W):* In this action, the robot changes its head orientation by randomly picking the value of the head pitch and head yaw from their allowable ranges.

*2) Look Towards Human (LTH):* During this operation, the robot remains sensitive, and in case a person is detected, the robot attempts to engage them by tracking them with its head.

*3) Wave hand (H):* During this action, the robot not only looks towards the people but also wave hand and says Hello.

*4) Handshake (HS):* For this action, the robot raises its right arm to a certain height and it waits at that position for three seconds for a handshake to happen. If the FSR touch sensor on the robot's right hand detects a touch, the robot grabs an object (hopefully the person's hand) that caused a touch, and utters a greeting phrase *nice to meet you*.

### 4.2. Event detector

The event detector provides the labels for three events, i.e., handshake, eye-contact and smile. The handshake label indicates if a touch has happened or not. This label is determined on the basis of touch sensor which is pasted on the robot's right hand (see Fig. 1). If a touch sensor detects a touch during the handshake action then it is considered that a handshake has established otherwise not.

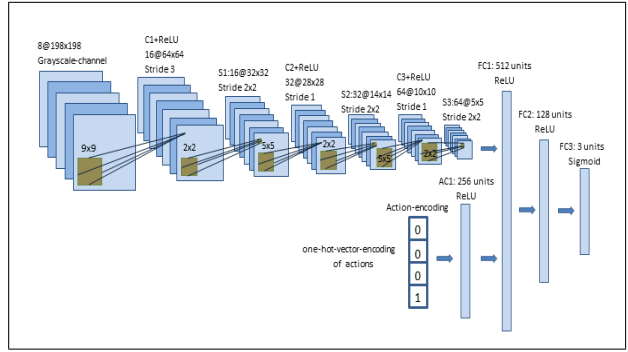

(a) Pnet: Deep Action-Conditional Prediction Network

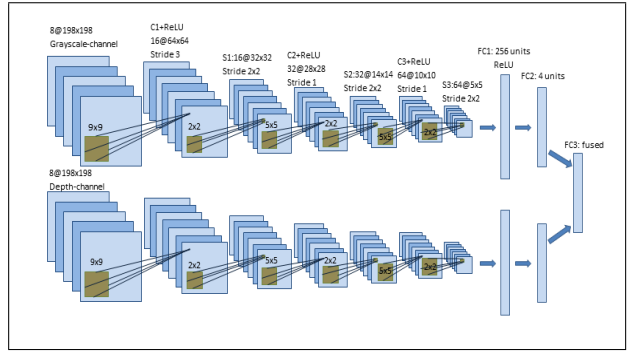

(b) Qnet: Deep Q-Network

Figure 4: Deep Neural Networks

The eye-contact and smile labels indicate if people looked towards the robot or not and smiled/laughed or not, respectively. These labels are given by the OpenCV based models[4] which take the eight most recent grayscale images as an input. The models scan through each of the eight input images and if the eye contact and smile events are present in any of the images, their labels are set to 1 otherwise 0.

### 4.3. Reward function g

The reward for Qnet is computed using the output of Pnet and the event detector. Pnet's output values are rounded off to one significant figure. These values are then compared with the labels from the event detector $e(t + 1)$ to determine the reward value. In the table 4, the reward function named "neutral" shows the reward formulation used in our proposed research, where # C indicates the number of correctly predicted events.

It can be seen that the agent gets a reward of value 1 and -0.1 if all of the three events are predicted correctly and incorrectly, respectively. A reward of value 0 is given if some of the events, but not all, are predicted correctly.

### 4.4. Pnet

Pnet comprises of two streams, i.e., grayscale-channel and action-encoding (see Fig. 4(a)). The input for the grayscale-channel is eight grayscale images while the input for the action-encoding stream is one-hot-vector encoding of an action.

The structure of Pnet is shown in Fig. 4(a). The grayscale-channel is the repetition of a sandwich, comprising of convolutional layer, non-linear ReLU and $2 \times 2$ max pooling of stride $2 \times 2$, till the first fully connected layer FC1. The layers C1, C2, C3 and FC1 convolve $16 \times 9 \times 9$ filters with a stride 3, $32 \times 5 \times 5$ filters with a stride 1 , $32 \times 5 \times 5$ filters with a stride 1, and $256 \times 5 \times 5$ filters with a stride 1, respectively. The FC1 layer has 512 units, 256 units are from preceding layer S3 while remaining 256 units are from the action encoding stream. The action-encoding stream transforms one-hot-vector encoding of an action into 256 units (AC1) which after applying ReLU is fed into FC1. This FC1 layer of Pnet, after applying ReLU, is further condensed to FC2 (128 units) which again after applying ReLU, is transformed to the layer FC3 (3 units). The output of FC3 is squashed between 0 and 1 by applying the Sigmoid function.

### 4.5. Qnet

Qnet comprises two identical streams (grayscale-channel and depth-channel) of deep neural networks (see Fig. 4(b)). The grayscale-channel and depth-channel take the eight most recent grayscale (G) and depth (D) images, respectively, to output the $q$-values for the four actions. Since the structure of two streams is identical, therefore, for brevity, we describe the model architecture of only one of the streams.

The structure of a Qnet stream is identical to the grayscale-channel of the Pnet till the layer FC1. However, unlike Pnet, the layer FC1 has only 256 units which after applying ReLU are further reduced to the four output units. These output units provide the $q$-values for each of the four actions, i.e., wait, look towards human, wave hand and handshake.

### 4.6. Training procedure

As mentioned in the section 3.3, our proposed training procedure comprises two phases, i.e., the data generation phase and the learning phase. Since, we exploit the dataset gathered in [13], therefore, the details of the data-generation phase are skipped. In [13], the robot interacted with people for 14 days, i.e., the number of episodes $M$ were 14 (see Algorithm 1). Each and every day, the data-generation phase was executed for around 4 hours followed by the execution of the learning phase. The exploration parameter $\epsilon$ was set to decay linearly from 1 to 0.1 over the 28,000 interaction steps. However, due to variations in the internet speed at different locations, the robot could only gather 13,938 interaction experiences during the 14 days of experimentation. These interaction experiences[5] were stored into the replay memory $\mathcal{M}$ for the off-policy training of our proposed neural models (Pnet and Qnet).

For the learning phase, the learning procedure (see Algorithm 1) was executed over the loop of 14 episodes. In each episode, a mini-buffer $\mathcal{B}$ of size 2000 interaction experiences were sampled from $\mathcal{M}$, and the mini-batch training of Pnet followed by Qnet was performed. The mini-batch $m$ could only hold up to 25 interaction experiences and the number of experience replays $n$ was set to

---

[5]The interaction experiences contained 111,504 grayscale and depth images as each state consists of eight most recent grayscale and depth frames.

| Events | Accuracy (%) |
| --- | --- |
| All | 88.9 |
| Handshake | 97.9 |
| Smile | 91.9 |
| Eye contact | 92.3 |

Table 1: Performance of Pnet.

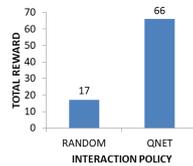

Figure 5: Qnet's ability to improve predictability

10.

Pnet was trained using the Stochastic Gradient Descent (SGD) method whereas Qnet's optimization was done using RMSprop (as suggested in [9]). The learning rate for both optimizers was kept constant at 0.00025. The discount factor for computing the objective function (equation 5) of the Qnet was also kept constant at 0.9. Finally, the old parameters of Qnet, i.e., $\hat{\boldsymbol{\theta}}$ were updated to the current parameters $\boldsymbol{\theta}$ after every episode.

## 5. Results

To evaluate the performances of Pnet and Qnet, we collected a test dataset. This test dataset comprises 560 interaction experiences which were gathered by the robot following a random policy. In addition, we also evaluate Qnet's behavior through a real world human-robot interaction experimentation. The remaining section summarizes the results of these evaluations.

### 5.1. Pnet ability to predict

We tested the performance of Pnet on a test dataset. The true labels for the test dataset were extracted using the event detector, and the results are summarized in the table 1. The first row indicates the overall accuracy, i.e., how often Pnet predicted all events (handshake, smile and eye contact) correctly, whereas the remaining rows show the accuracy of predicting the individual events correctly.

### 5.2. Qnet ability to improve predictability

To evaluate Qnet's behavior before and after learning, we conducted a three days of human-robot interaction experimentation in a public place. Each day the robot executed 600 interaction steps for around 2 hours during the lunch break. On the first day, the robot was given a random policy, i.e., it selected the actions randomly while on the second day, the robot used the trained Qnet policy for the interaction. We noted the total reward accumulated following the reward function definition (see section 4.3), and the results are presented in Fig. 5. It can be seen that Qnet gathered the total reward of value 66 while the random policy could only accumulate 17 reward values. This result validates that Qnet has a behavior which improves the predictability of Pnet to maximize the reward for itself. The third day of experimentation was done using the model [13], and its findings are highlighted in the next section.

### 5.3. Similarity of Qnet's behavior to the human social behavior

To evaluate the similarity between Qnet's behavior and the human social behavior, we employed the following evaluation procedure. The three volunteers evaluated the decisions of the Qnet on a test dataset. Each volunteer observed the sequence of eight grayscale frames depicting the scenario followed by the Qnet's decision (the action with maximum $q$-value). The volunteer then decided if the decision is in agreement with what human would do in that situation. If the majority of volunteers considered the decision to be inappropriate, then they were asked to pick up the most appropriate action for the depicted scenario. By following this evaluation procedure, we evaluated several policy networks as shown in the table 2 and 3.

| Trained Model | Qnet$_d$ | | | Qnet$_{im}$ | | | Qnet$_{rand}$ |
|---|---|---|---|---|---|---|---|
| Reward type | Direct | | | Intrinsic | | | - |
| State input | G-D | G | D | G-D | G | D | - |
| Overall | 0.949 | 0.691 | 0.682 | 0.968 | 0.934 | 0.914 | 0.238 |
| W | 0.921 | 0.517 | 0.454 | 0.925 | 0.905 | 0.781 | 0.160 |
| LTH | 0.972 | 0.541 | 0.635 | 0.994 | 0.993 | 0.930 | 0.222 |
| H | 0.812 | 0.478 | 0.348 | 0.967 | 0.906 | 0.935 | 0.186 |
| HS | 0.967 | 0.106 | 0.781 | 0.968 | 0.923 | 0.946 | 0.328 |
| Handshake ratio | 0.481 | - | - | 0.514 | - | - | 0.170 |

Table 2: Similarity of the learned behavior to the human social behavior via F1-scores.

| Model | Qnet$_d$ | | | | | Qnet$_{im}$ | | | | |
|---|---|---|---|---|---|---|---|---|---|---|
| Reward type | Direct | | | | | Intrinsic | | | | |
| State Input | G-D | | | | | G-D | | | | |
| Measure (%) | TPR | TNR | FPR | FNR | Acc | TPR | TNR | FPR | FNR | Acc |
| Overall | 94.8 | 98.3 | 1.70 | 5.12 | 97.4 | 96.8 | 98.9 | 1.07 | 3.21 | 98.4 |
| W | 100.0 | 98.7 | 1.30 | 0.00 | 98.8 | 86 | 100.0 | 0.00 | 14.0 | 97.5 |
| LTH | 96.9 | 99.4 | 0.56 | 3.15 | 98.9 | 99.4 | 99.8 | 0.25 | 0.63 | 99.6 |
| H | 69.2 | 99.8 | 0.17 | 30.8 | 96.2 | 96.1 | 99.6 | 0.41 | 3.95 | 99.1 |
| HS | 98.5 | 91.3 | 8.73 | 1.46 | 95.8 | 100.0 | 95.5 | 4.49 | 0.00 | 97.3 |

Table 3: Confusion table of the learned human-like behavior policies.

In the table 2 and 3, the first three rows provide the specifications of the trained models. The Qnet$_d$ correspond to the model in [13] which is trained by giving it a direct reward for maximizing the number of successful handshakes. Therefore, the reward type for Qnet$_d$ is indicated as direct. The definition for the direct rewarding is as follows. Qnet$_d$ gets the reward of 1 and -0.1 on a successful and unsuccessful handshake, respectively, and a reward of value 0 for actions other than a handshake. The handshake is considered to be successful/unsuccessful if touch sensor detects/does-not-detect the touch in response to robot's handshake action. The model Qnet$_{im}$ corresponds to our proposed model which has an intrinsically motivated learning objective. Hence, the reward type for $Qnet_{im}$ is mentioned as intrinsic. The model $Qnet_{rand}$ corresponds to a random policy in which the actions where selected randomly from the action space. Finally, under model specifications, the row labeled state input tells which stream of the dual stream convolutional neural networks was evaluated. The $Qnet_{im}$ and $Qnet_d$ comprise two stream of convolutional neural networks, i.e., grayscale (G) stream and depth (D) stream. In addition to the evaluation of $Qnet_{im}$ and $Qnet_d$, we also evaluated the performance of their individual streams.

The remainder of the table 2, except last row, indicates the likeness of the indicated models to the human behavior in terms of F1-scores. The F1 score measures the accuracy of a statistical model where F1-score=1 and F1-score=0 indicate that the model performance is best and worst, respectively. It can be seen in the table 2 that the overall performance, i.e., the likeness to human behavior, of $Qnet_{im}$ is better than $Qnet_d$. The performance in selecting individual actions by $Qnet_{im}$ is also significantly higher than that of $Qnet_d$. Furthermore, we also evaluated the individual streams of both $Qnet_{im}$ and $Qnet_d$. It can be seen that the grayscale (G) and depth (D) streams of $Qnet_{im}$ gave superior performances compared to the streams of $Qnet_d$.

The last row of table 2 mentions the hand-shake ratio. This ratio corresponds to the ratio of the number of successful handshakes performed by the robot to the total number of handshake attempts. These handshake ratios for $Qnet_{im}$, $Qnet_d$ and $Qnet_{rand}$ were calculated during the experimentation described in section 5.2. It can be seen that the handshake ratio of $Qnet_{im}$ is similar to $Qnet_d$. Although, unlike $Qnet_d$, the objective of $Qnet_{im}$ was not to maximize the number of handshakes. However, by exploiting the predictability of human behavior, $Qnet_{im}$ successfully learned to select the appropriate actions in every given scenario.

Table 3 further compares the best models, from table 2, of the two methods, i.e., direct rewarding method ($Qnet_d(G-D)$) and intrinsic motivation method ($Qnet_{im}(G-D)$). The comparison is made in terms of True Positive Rate (TPR), True Negative Rate (TNR), False Positive Rate (FPR), False Negative Rate (FNR) and Accuracy (Acc). It can be seen that the $Qnet_{im}$ demonstrated superior performance compared to $Qnet_{d}$ in terms of all performance measure metrics.

It should also be noted that the models $Qnet_{im}$ and $Qnet_{d}$ are compared to highlight two points. First, the human behavior is predictable, and the predictive model based intrinsically motivated learning leads to more human-like behavior than traditional direct reward based learning methods like [13]. Second, the reward is usually sparse in the real-world HRI problems thus, intrinsically motivated learning is beneficial to ensure human-like behavior.

Figs.6-9 show the instances from the test dataset in which the $Qnet_{im}$ made the correct decisions, i.e., the decisions which were in accordance to the decision of human evaluators (HE). The robot learned to take the waiting (W) action when there were no people around the robot, people were busy in some activity or people were going away from the robot (see Fig. 6). The look towards human (LTH) action, as shown in Fig. 7, was taken when the people were mildly engaged with the robot. The term mildly engaged refers to the engagement in which either people were gazing at the robot from distance or people were in front of the robot but they were not looking at it. The wave hand (H) action (see Fig. 8) was taken when people were distant and they were not looking at the robot. Lastly, the handshake (HS) action, as shown in Fig. 9, was taken when people were fully engaged with the robot. The full engagement means the people were not only standing in front of the robot but also looking at it. The robot's successful decisions and their implications are further discussed in detail in the next section.

Finally, Fig. 10 shows the scenarios from the test dataset in which $Qnet_{im}$ made the incorrect decision, i.e., there was a contradiction in the decision of human evaluators (HE) and $Qnet_{im}$.

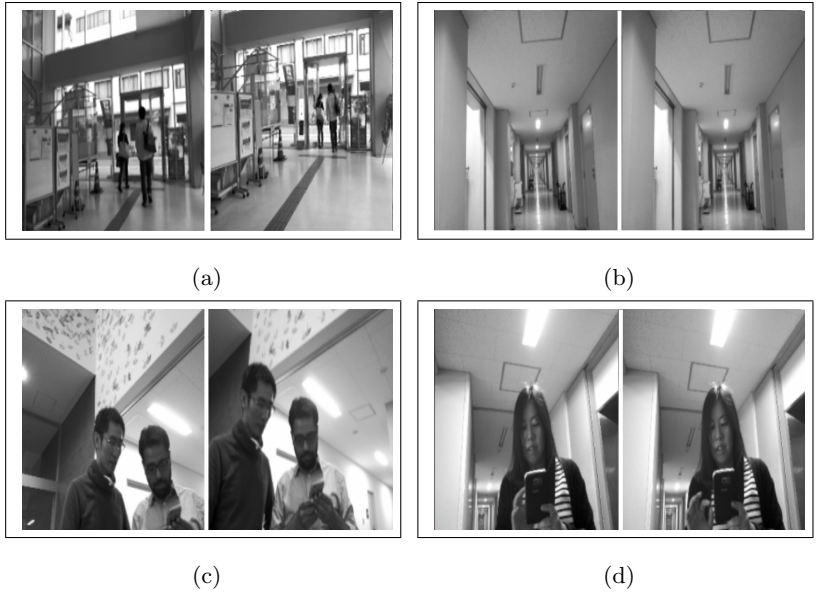

Figure 6: Example scenarios in which a **Wait** action was choosen.

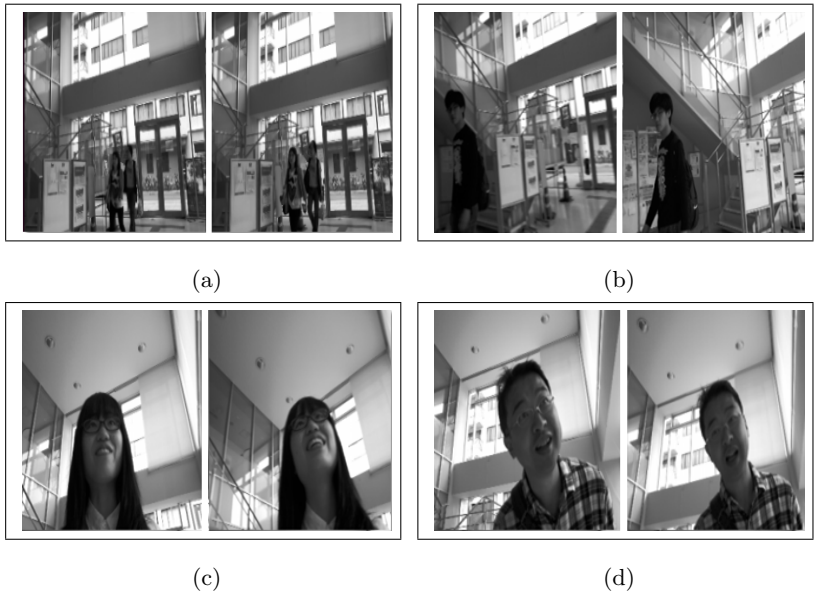

Figure 7: Example scenarios in which a **LookTowardsHuman** action was choosen.

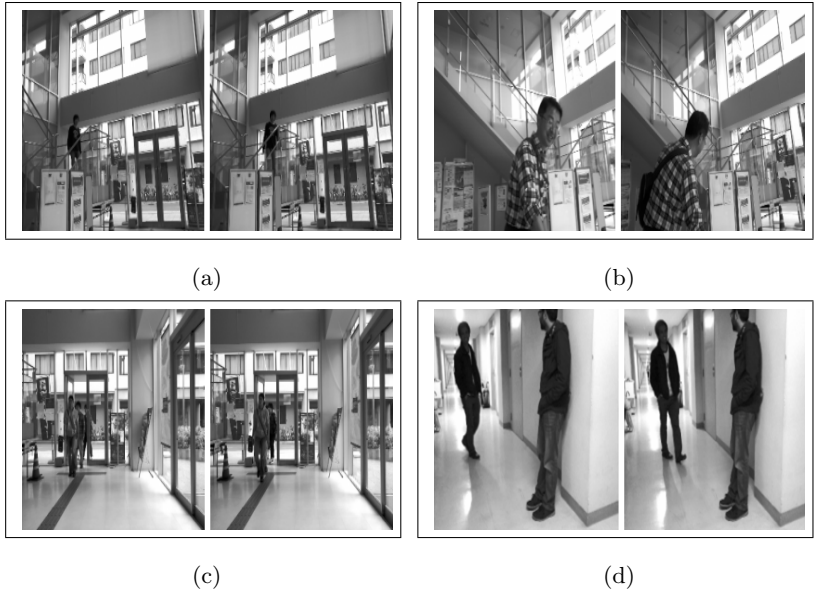

(a)                              (b)

(c)                              (d)

Figure 8: Example scenarios in which a **WaveHand** action was choosen.

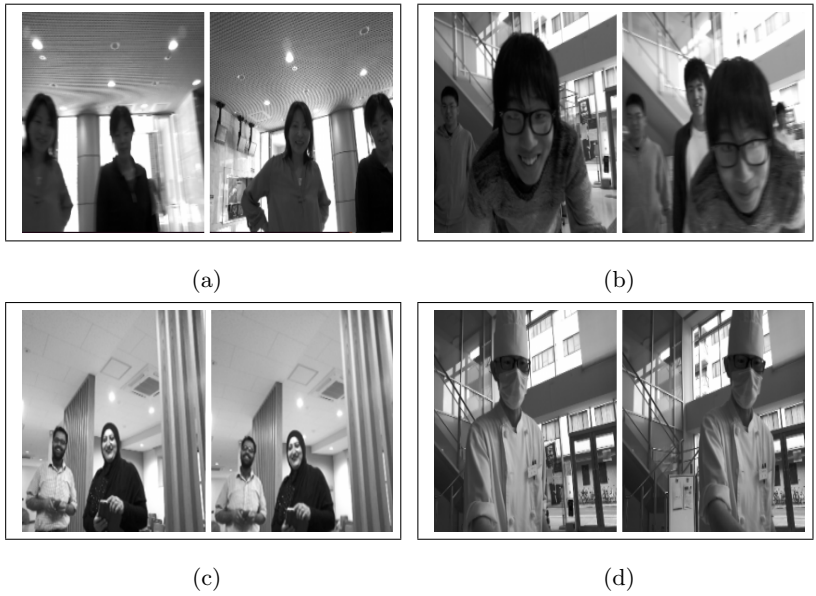

(a)                              (b)

(c)                              (d)

Figure 9: Example scenarios in which a **Handshake** action was choosen.

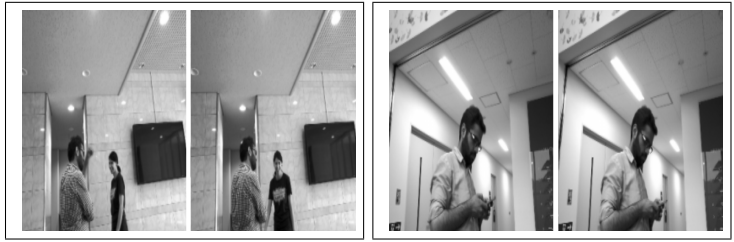

(a) Qnet$_{im}$: Wait ; HE: Wavehand          (b) Qnet$_{im}$: LTH ; HE: Wait

Figure 10: Unsuccessful cases of Qnet$_{im}$ decisions

## 6. Discussion

This section discusses the i) human-like behavior of our presented polices, ii) ability of our proposed Qnet$_{im}$ to recognize human intentions; iii) reward function, its physical meaning and the impact of different reward functions on the robot behavior; iv) the limitation of the current system. For brevity, in this section, the Qnet$_{im}$ is referred as Qnet.

### 6.1. Likeness to human behavior

Note that the objective of $Qnet_{im}$ and $Qnet_d$ was to improve the predictability of human behavior and to maximize the number of successful handshakes, respectively. We evaluate our policies ($Qnet_{im}$ and $Qnet_d$) for likeness-to-human-behavior/social-acceptance to highlight that our learned policies exhibit human-like social behavior even though their objectives functions were not meant to maximize the social acceptability. Furthermore, the reason for not giving a reward/penalty directly on the basis of human-likeness/social-acceptance is that the human actions show a tremendous diversity and thus, it is not possible to quantify the notion "socially acceptable behavior" to define a reward metric. Therefore, we present alternative reward function that implicitly inculcates human-like behavior into the optimal policy for social human-robot interaction.

### 6.2. Intention recognition

The instances shown in the Figs. 6-9 and the Qnet's decision in these scenarios insinuate that the proposed Qnet has an understanding of the intention depicting factors, like people walking trajectory, people level of engagement with the robot, head orientation, activity in progress, etc, and their implications. In human-human interaction, these factors help other people to choose the appropriate action for the interaction. Likewise, Qnet's understanding of these factors further validates the high similarity between Qnet's behavior and the human social behavior.

In Figs. 6(a), 7(a) and 8(c) the first and last frames out of eight most recent frames are shown to indicate the people walking trajectory. In the Fig. 6(a), a person is walking away from the robot and thus, Qnet decides to wait. However, in Figs. 7(a) and 8(c), people are approaching towards the robot so Qnet decided to look towards them and wave hand, respectively, to get their attention for the interaction.

The understanding about the level of human engagement with the robot can be seen from Figs. 7-9. In Fig. 7, either people are standing close to the robot or people are at a small distance away but they are looking at the robot.

Therefore, Qnet chooses the look towards human action which is a softer way of gaining people's attention. In Fig. 8, people are, relatively, at a longer distance, thus the level of engagement is lower compared to the scenarios shown in Fig. 7. Therefore, in Fig. 8 scenes, Qnet picked a wave hand action which is a stronger way of getting the people's attention. Finally, in Fig. 9, people are fully engaged with the robot thus the agent picks the handshake action.

The agent has also learned that if people are not around the robot then no action other than waiting will lead to a reward (see Fig. 6(b)). Finally, Qnet's understanding about the ongoing activities is evident from its action in situations indicated in Figs 6(c) and 6(d). In Fig. 6(c), a discussion between two people, standing in front of the robot, is going on and the agent decides to wait. Likewise, in Fig. 6(d), the person is taking robot's picture and therefore, Qnet decides to wait.

### 6.3. Reward function

This section discusses the different aspects of the proposed reward function. The first section describes the physical meaning, i.e., the intuition behind the the formulation intrinsic motivation. The second section explains the reason for choosing particular values for the rewards and penalties.

### 6.3.1. Physical meaning

The reward for Qnet is given as to improve the action-conditional predictability of Pnet. In our proposed method, Pnet predicts three events, i.e., handshake, eye-contact and smile. In social human-robot interaction, these three events can be seen as the indicators of the good versus bad human-robot interactions. For instance, if a robot chooses an action that entices a human to do smile or perform a handshake with a robot, then it is assumed that a good human-robot interaction has occurred. On the other hand, if a robot chooses an action that does not instantiate any of the three events, the interaction is assumed to be banal/bad.

| # C | 0 | 1 | 2 | 3 |
|---|---|---|---|---|
| Strict | -0.1 | -0.1 | -0.1 | 1.0 |
| Neutral | -0.1 | 0.0 | 0.0 | 1.0 |
| Kind | -0.1 | 0.8 | 0.9 | 1.0 |

Table 4: Rewards functions

The reward function for Qnet is designed to instill an ability within Qnet to foresee the occurrence of the above-mentioned events in the next time-step $t+1$ in response to its decision at time-step $t$. This implies that Qnet predicts either the good social interaction will occur or not and assigns the highest $q$-value to the action through which it is most certain about the future. In other words, Qnet selects the action for which it can confidently foresee the consequences. This kind of behavior is very similar to human behaviors during their social interactions. As human unconsciously or consciously prefer the actions for which they can envisage the response of their interacting human partners [31].

### 6.3.2. Impact on Qnet's behavior

All of the results presented so far were related to the reward function which inculcated the most socially acceptable behavior into the robot. In this section, we investigate the impact of various other reward formulations on the Qnet's behavior. The same evaluation procedure as for Table 2, i.e., measuring similarity to human behavior, was followed to evaluate the Qnet's behavior in response to its rewarding. Fig 11(a) shows the impact of three different reward functions named strict, neutral and kind on the Qnet's behavior. Table 4 formalizes these three reward functions, where # C indicates the number of correct predictions done by Pnet.

In all of the three reward functions, Qnet gets a positive reward of 1 and a negative reward of 0.1 on the successful and unsuccessful prediction of all events by Pnet, respectively. However, these reward functions differ from each other in term of the reward given on the correct prediction of some of the events but not all. In Fig. 11(a), it can be seen that the reward function named neutral inculcates the most acceptable social behavior into the robot.

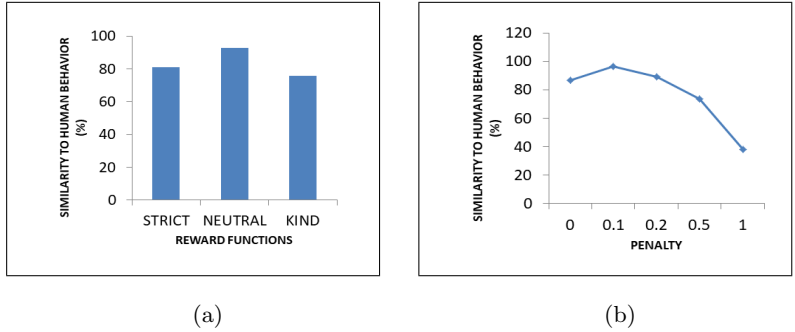

(a)                           (b)

Figure 11: Effect of reward function on the robot's behavior.

To further investigate the reward function named neutral, we trained five more models by varying the penalty on an incorrect prediction of all events by the Pnet. The rest of the reward function was kept same, i.e., on correct prediction of all events and correct prediction of some of the events but not all, Qnet received the reward of 1 and 0, respectively. The penalties for the five models on the incorrect prediction of all events were 0, 0.1, 0.2, 0.5 and 1. Fig. 11(b) presents the results, and it shows that the reward function with 0.1 penalty gave the most acceptable social behavior to the robot.

It should be noted that the reward functions with lower penalties (negative reward) on incorrect predictions demonstrate the most socially acceptable behavior. The reason is that the strict reward functions which gave high penalties (e.g., -1) on incorrect prediction of events made Qnet reluctant to use wave hand and handshake actions for the interaction. Since, these two actions can strongly alter the human on-going behavior, therefore, the prediction task can become harder. It was observed that with high penalties, instead of learning to improve predictions, Qnet hacked the reward function by simply taking the wait action and rarely choosing the look towards human action. Hence, the strict reward functions lead to unacceptable social behaviors and more acceptable behavior emerged with lower penalties.

*6.4. Limitation of the current system*

The current system lacks memory and has a limited action space. The lack of memory makes the robot oblivious of what actions it has already executed with the interacting partner. This causes the robot to exhibit repetitive behavior. For instance, it has been observed that if a person, after shaking hand with a robot, continues to stand in the same posture, the agent selected the handshake action again. In our daily life, people are not used to this kind of repeating interaction behavior. Therefore, it was observed that after few seconds of interaction, the people stopped responding to the robot actions. This limitation can also be observed in the results given in Fig. 5 and the handshake ratio given in Table 2. In Fig. 5, the robot executed 600 interaction but couldn't gather much reward. Likewise, the handshake ratio in Table 2 is also not high. The possible solution to this limitation has been made the part of our future work, i.e., augmenting policy with a memory and increasing the action space. The memory will enable the policy to keep track of person's identity (e.g., face recognition based identity) and thus, the repeating behavior can be easily avoided. On the other hand, the enhanced action space will mitigate the issue of seeing robot behavior as repetitive.

## 7. Conclusion

For the robots to enter our social world, it is important for them to learn the human-like social behavior through interaction with people in the real reward-sparse environments. In this paper, we propose the intrinsically motivated deep reinforcement learning framework for learning in reward-sparse real world. The intrinsic motivation is inculcated by our novel deep action-conditional predictor using which the deep reinforcement learning agent gets higher rewards on the accurate prediction of the future states. By using this framework, the robot learned human-like social interaction skills through off-policy learning on the 14 days of real-world interaction experiences collected in [13]. The results indicate that the i) robot has learned to respond to the complex human interactive behaviors with the utmost propriety, ii) robot interprets the human behavior by intention depicting factors (e.g., human body language, walking trajectory or any ongoing activity, etc.), and iii) intrinsically motivated reinforcement learning framework learned more human-like behavior than usual reinforcement learning method in the real world where the direct reward is not always available

In our future work, we plan to make our robot system mobile so that it can actively approach people for safe, natural and effective social interaction. To do so, we plan to propose a framework similar in notion to active object recognition method [32] where the neural model actively learns camera positioning to accomplish object recognition task using extreme trust region policy optimization. Another future milestone is to augment our model with neural memory networks for learning a sequential long-term human-robot interaction. In addition, we also plan to increase the robot action space instead of restricting it to four actions only.

**Acknowledgment**

This work was supported by JST ERATO Grant Number JPMJER1401, Japan